\documentclass[wcp]{jmlr}

\usepackage{longtable}

\usepackage{natbib}

\usepackage{threeparttable}
\usepackage{multirow}
\usepackage{floatrow}
\usepackage{pifont}

\makeatletter
\let\Ginclude@graphics\@org@Ginclude@graphics
\makeatother

\jmlrvolume{}

\title[Open Images V5 Text Annotation and YAMTS]{Open Images V5 Text Annotation and Yet Another Mask Text Spotter}

 \author{\Name{Ilya Krylov} \Email{ilya.krylov@intel.com}\\
  \addr IOTG Computer Vision (ICV), Intel
  \AND
  \Name{Sergei Nosov} \Email{sergei.nosov@intel.com}\\
  \addr IOTG Computer Vision (ICV), Intel
  \AND
  \Name{Vladislav Sovrasov} \Email{sovrasov.vladislav@itmm.unn.ru}\\
  \addr Lobachevsky State University of Nizhny Novgorod, Russia \\
        IOTG Computer Vision (ICV), Intel
  }

\begin{document}

\maketitle

\begin{abstract}
  A large scale human-labeled dataset plays an important role in creating high quality deep learning models. In this paper we present text annotation for Open Images V5 dataset. To our knowledge it is the largest among publicly available manually created text annotations. Having this annotation we trained a simple Mask-RCNN-based network, referred as Yet Another Mask Text Spotter (YAMTS), which achieves competitive performance or even outperforms current state-of-the-art approaches in some cases on ICDAR~2013, ICDAR~2015 and {Total-Text} datasets. Code for text spotting model available online at: \url{https://github.com/openvinotoolkit/training_extensions}. The model can be exported to OpenVINO{\texttrademark}-format and run on Intel{\textregistered} CPUs.
\end{abstract}
\begin{keywords}
Scene text dataset, Text spotting dataset, Scene text detection and recognition, Text spotting
\end{keywords}

\section{Introduction}

Text spotting has attracted many researchers since its practical use has a lot of applications such as instant translation, media retrieval, license plate recognition, video tagging, assistance for blind and visually impaired people. Text spotting includes simultaneous text detection and recognition.

Despite tremendous progress has been made recently (\cite{crafts, mango, abcnet, MTSv3}),  text spotting is still unsolved and challenging task complicated by various factors such as arbitrary background, extreme aspect ratios, and diverse shapes of scene text instances. In order to train an accurate model which will be able to tackle aforementioned problems it is crucial to have a large scale annotated dataset.

In this paper we present the largest to our knowledge publicly available human-labeled annotation containing precise localization of text instances together with their transcriptions.

Our experiments show that having this annotation together with corresponding images as a part of training set makes it feasible to train simple Mask-RCNN-based (\cite{he2017}) model without any specific tricks that achieves competitive performance on such popular text spotting benchmarks as ICDAR~2013, ICDAR~2015 and Total-Text.

\section{Related Works}

In this section we consider publicly available scene text spotting datasets and approaches to tackle the text spotting problem.

\subsection{Scene Text Datasets}
Here, we briefly describe publicly available scene text datasets that are often used by researchers.

\textbf{SynthText} (\cite{gupta2016synthetic}) is widely-used synthetically generated dataset, containing approximately 800K images and 6M synthetic text instances. Each text instance is annotated with its text-string, word-level and character-level bounding boxes.

\textbf{UnrealText} (\cite{long2020unrealtext}) contains 600K synthetic images with 12M cropped text instances. The synthesis engine is implemented based on Unreal Engine and the UnrealCV plugin.

\textbf{Street View Text (SVT)} (\cite{svt}) dataset was harvested from Google Street View. Image text in this data exhibits high variability and often has low resolution. It contains 100 images for training and 250 for testing.

\textbf{ICDAR~2003} (\cite{icdar03}) contains 509 natural scene images (including 258 training images and 251 test images) in total.

\textbf{ICDAR~2013} (\cite{icdar13}) consists of 229 training images and 233 testing images, with word-level annotations provided. It is the standard benchmark dataset for evaluating near-horizontal text detection. In addition, a set of vocabularies are provided:
\begin{itemize}
\item Strongly Contextualised: per-image vocabularies of
100 words comprising all words in the corresponding
image as well as distractor words selected from the
rest of the training/test set.
\item Weakly Contextualised: a vocabulary of all words in
the training/test set.
\item Generic: a generic vocabulary of about 90K words.
\end{itemize}

\textbf{ICDAR~2015} (\cite{icdar15}) contains 1,500 images: 1,000 for training and 500 for testing. ICDAR~2015 refers to images where text captured incidentally. As in case of ICDAR~2013, a set of vocabularies are provided: Strongly Contextualised, Weakly Contextualised, Generic.

\textbf{ICDAR~2017 MLT} (\cite{icdar17mlt}) contains 18,000 images in total: 7,200 images refer to training subset, 1,800 refer to validation subset and 9,000 refer to test subset. The dataset is composed of complete scene images which come from 9 languages representing 6 different scripts. The number of images per script is equal. This makes it a useful benchmark for the task of multi-lingual scene text detection. The considered languages are the following: Chinese, Japanese, Korean, English, French, Arabic, Italian, German and Indian.

\textbf{ICDAR~2019 MLT} (\cite{icdar19mlt}) consists of 10,000 images in training set and 10,000 in test set. ICDAR~2019 MLT is also multi-lingual dataset and it includes ten languages, representing seven different scripts: Arabic, Bangla, Chinese, Devanagari, English, French, German, Italian, Japanese, and Korean. The number of images per script is equal.

\textbf {SCUT-CTW1500} (\cite{scutctw1500}) dataset contains 1,500 images: 1,000 for training and 500 for testing. The images are manually harvested from the Internet, image libraries such as Google Open-Image, or phone cameras. The dataset contains a lot of horizontal and multi-oriented text.

\textbf{Total-Text} (\cite{totaltext}) is a text detection dataset that consists of 1,555 images with a variety of text types including horizontal, multi-oriented, and curved text instances. The training split and testing split have 1,255 images and 300 images, respectively.

\textbf{ICDAR~2019 ART} (\cite{icdar19art}) dataset is a combination of Total-Text, SCUT-CTW1500 and Baidu Curved Scene Text, which were collected with the motive of introducing the arbitrary-shaped text problem to the scene text community. On top of the existing images (3,055), more than 7,111 images are added to mixture of both datasets, which make ICDAR~2019 ART one of the larger scale scene text datasets today. There is a total of 10,166 images: a training set with 5603 images and a testing set of 4563 images.

\textbf{COCO-Text} (\cite{cocotext}) is a large-scale scene text dataset, based on MS COCO. It contains 63,686 images with 239,506 annotated text instances.

\textbf{TextOCR} (\cite{textocr}) is a recently published arbitrary-shaped scene text detection and recognition dataset consisting of 28,134 images with 900K annotated words.

\subsection{Text spotting}

\subsubsection{Text detection and recognition methods}

Text detection and recognition approaches carry out text spotting via detecting each text instance using a text detection model and then recognizing the cropped text instance by a separate text recognition model. Text detection model can be represented by a polygon detection model or by an instance segmentation model.

An example of such approach is \textbf{TextBoxes++} (\cite{textboxespp}) which is inspired by SSD (\cite{ssd}) model, it relies on specifically designed text-box layers to detect text instances having extreme aspect ratios. TextBoxes++ can generate arbitrarily oriented bounding boxes in terms of oriented rectangles or general quadrilaterals to deal with oriented text. The combination of TextBoxes++ and CRNN (\cite{crnn}) recognizer achieves favorable results in text spotting.

Major drawback of such methods is that the propagation of error between the detection and recognition models does not happen that leads to worse performance than in case of end-to-end methods.

\subsubsection{End-to-end methods}

End-to-end approaches do text spotting by a single holistic model that performs text detection and recognition simultaneously. These methods have become dominating and currently show better results, benefiting from joint text detection and recognition training.

\textbf{ABCNet} (\cite{abcnet}) is a single-shot, anchor-free detector-based text spotter. It introduces concise parametric representation of curved scene text using Bezier curves. BezierAlign is proposed for accurate feature alignment so that the recognition branch can be naturally connected to the overall structure.

\textbf{Mask TextSpotter v3} (\cite{MTSv3}) is an end-to-end arbitrary-shape scene text spotter. Instead of using RPN it introduces SPN to generate proposals, represented by accurate polygons. Hard RoI masking is directly applied to the extracted RoI features to suppress background noise or neighboring text instances.

\textbf{CRAFTS} (\cite{crafts}) proposes an end-to-end network that could detect and recognize arbitrary-shaped texts by tightly coupling detection and recognition modules. Spatial character information obtained from the detector is used in the rectification and recognition module.

\textbf{MANGO} (\cite{mango}) is a one-stage scene text spotter. This model removes the RoI operations and designs the position-aware attention module to
coarsely localize the text sequences. After that, a lightweight
sequence decoder is applied to obtain all of the final character sequences into a batch.

\section{Open Images V5 Text Annotation}

Open Images V5 dataset contains about 9 million varied images. The images are very diverse and often contain complex scenes with several objects. Typically text instances appear on images of indoor and outdoor scenes as well as artificially created images such as posters and others. The dataset consists of following subsets: training (16 archives such as train\_0, ..., train\_f), validation, test. We annotated several archives of Open Images V5 namely: train\_1, train\_2, train\_5, train\_f, validation. The annotation has been made using Computer Vision Annotation Tool (CVAT), it is available at: \url{https://github.com/openvinotoolkit/cvat}.

\subsection{Annotation Details}
A text instance is an uninterrupted sequence of characters separated by a space. Text instances are annotated by polygons with sufficient number of points to make a tight contour. Only English letters and numbers are annotated. Text instances have Boolean attributes: machine-printed, legible. Images without English (or other language where we can use the same letters to annotate) text are not annotated. Images with many instances, such as newspapers, book pages and etc. are also skipped.

As a result, we have 207K images, 2.5M text instances and 2M of them have annotated transcription.

The annotation is stored in MS COCO-like format, where each annotated text instance has additional \texttt{attributes} field along with \texttt{bbox}, \texttt{segmentation} and other. The \texttt{attributes} field has the following sub-fields: \texttt{transcription} - string, \texttt{legible} - Boolean, \texttt{machine\_printed} - Boolean. The dataset statistics are shown in Table~\ref{table:stat} and comparison with some popular datasets with know number of text instances is shown in Table~\ref{table:comp}.

The annotation is available at: \url{https://storage.openvinotoolkit.org/repositories/openvino_training_extensions/datasets/open_images_v5_text}.

\begin{table}[ht]
\centering 
\begin{threeparttable}
\caption{Datasets statistics.} 
\begin{tabular*}{\textwidth}{l @{\extracolsep{\fill}} ccc} 
\hline 
Subset & Images & Instances & Legible \\
\hline 
train\_1 & 50,878 & 540,057 & 444,001 \\
train\_2 & 48,877 & 594,836 & 503,171 \\
train\_5 & 49,329 & 621,057 & 496,203 \\
train\_f & 41,975 & 597,352 & 471,050 \\
validation & 16,731 & 218,308 & 158,962 \\ [1ex] 
\hline
TOTAL & 207,790 & 2,571,610 & 2,073,387 \\ [1ex] 
\hline 
\end{tabular*}
\label{table:stat} 
\end{threeparttable}
\end{table}

\begin{table}[ht]
\centering 
\begin{threeparttable}
\caption{Comparison with datasets.} 
\begin{tabular*}{\textwidth}{l @{\extracolsep{\fill}} cc} 
\hline 
Dataset & Images & Text instances \\
\hline 
ICDAR~2013 & 462 & 1,944 \\
ICDAR~2015 & 1,500 & 17,116 \\
SCUT-CTW1500 & 1,500 & 10,751 \\
Total-Text & 1,555 & 11,459 \\
COCO-Text & 63,686 & 239,506 \\
TextOCR & 28,134 & 903,069 \\ [1ex] 
\hline
Open Images V5 Text & 207,790 & 2,571,610 \\ [1ex] 
\hline 
\end{tabular*}
\label{table:comp} 
\end{threeparttable}
\end{table}

\subsection{Evaluation Protocol}
As the evaluation protocol of end-to-end recognition we follow ICDAR~2015 end-to-end recognition protocol while changing the representation of polygons from four vertexes to an arbitrary number of vertexes in order to handle the polygons of arbitrary shapes. This evaluation protocol is widely-used for Total-Text end-to-end evaluation (e.g. \cite{MTSv2}, \cite{MTSv3}, \cite{mango}).

\section{Yet Another Mask Text Spotter}

\subsection{Methodology}

\begin{figure}[H]

\caption{Overall architecture of YAMTS. The detection branch provides bounding boxes that play role of RoIs for mask and text branches. The mask branch generates word segmentation, the text recognition branch encodes extracted features even more, then GRU-based decoder takes previously generated symbol and attention-applied encoder outputs and generates next symbol until the End-Of-Sequence symbol is met.}
\vspace{1em}
\centering
\includegraphics[scale=0.284]{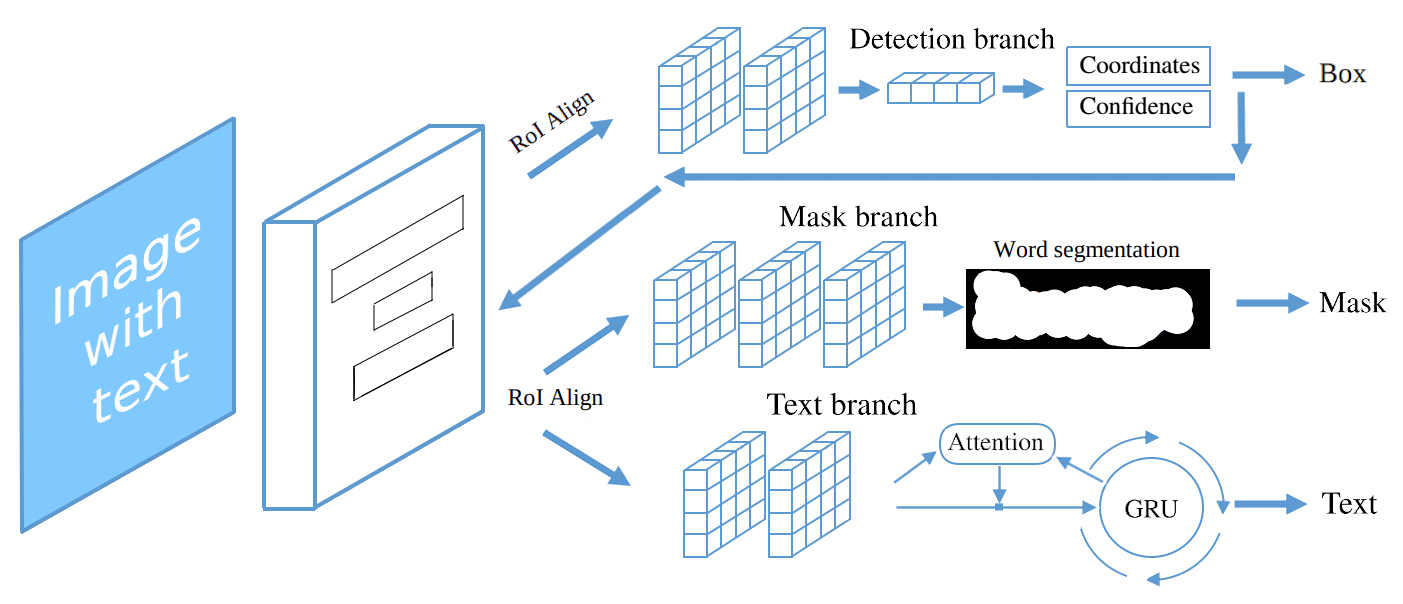}
\label{figure:figure1} 
\end{figure}

We adopt Mask-RCNN as our text detection branch with ResNet-50 followed by FPN. A text recognition head has been added in addition to detection and mask head. The text recognition head is presented by encoder, decoder and additive spatial attention, see Figure~\ref{figure:figure1}.

A fixed-size input (28x28) to the text recognition encoder is produced by RoIAlign layer. The text recognition encoder is represented by 3 stacked convolution layers with 3x3 kernel size. Each of them is followed by batch normalization and ReLU activation. As in \cite{MTSv2} we use spatial Bahdanau attention mechanism (\cite {bahdanau2014neural}) to prepare an input to the text recognition decoder. The text recognition decoder is based on GRU with 256-dimensional hidden state. Attention weights are computed at each decoding step using the text recognition encoder's outputs and current hidden state of the decoder. These weights are multiplied by the encoder outputs to create a weighted combination. The weighted combination and previously generated symbol is used by the decoder to produce new one until the End-Of-Sequence symbol is met.

\subsection{Datasets}

Datasets used in training are listed in this section. We do not use any synthetic dataset for pre-training. Also we do not use any special sampling technique to regulate frequency of getting images of some particular dataset, we just combine all the following datasets into a single one and use it for training:
\begin{itemize}
    \item Open Images V5 training subsets: 1, 2, 5, f.
    \item ICDAR~2013 training subset.
    \item ICDAR~2015 training subset.
    \item ICDAR~2017 MLT training and validation subsets, images with Latin text only.
    \item ICDAR~2019 MLT training subset, images with Latin text only.
    \item ICDAR~2019 ART training subset with excluded images related to Total-Text dataset.
    \item COCO-text training  and validation subsets.
    \item MSRA-TD500 training and test subsets.
\end{itemize}

As a result, we have 234K images in total with about 2.8M text instances and 2.2M of them have transcription, that means that they are suitable for text recognition head training.

\subsection{Implementation details}

The model is trained using 4 GPUs with 2 images per GPU as a batch size during 150 epochs. The backbone of the model is initialized by ImageNet (\cite{imagenet}) pre-trained weights. Data augmentation includes multi-scale training, pixel-level augmentations, and arbitrary-angle rotations in range of [-90$^{\circ}$, 90$^{\circ}$]. The model is optimized using SGD with a weight decay of 0.0001 and a momentum of 0.9. The Cosine Annealing is used as a learning rate schedule. The negative log likelihood loss is used to the text recognition head.

At the test time, an input image is resized to 1280x768 without keeping aspect ratio in case of ICDAR~2013, ICDAR~2015, Open Images V5 datasets. {In case of Total-Text dataset, images are resized keeping aspect ratio since there is a significant number of vertical images}. We also use 1600x960 input resolution on ICDAR~2015 since it contains a lot of small text instances.

When performing inference with a lexicon, a weighted edit distance algorithm as proposed in \cite{MTSv2} is used to find the best matching word. The weights of weighted edit distance depend on the character probability which is yielded by the text recognition decoder.

The model is implemented so that it can be easily exported to OpenVINO{\texttrademark}-format and run on Intel{\textregistered} CPUs. See details at: \url{https://github.com/openvinotoolkit/training_extensions}.

\subsection{Experiments}

In order to validate the effectiveness of YAMTS, we conduct experiments and compare with other state-of-the-art
methods on three most frequently used in evalutaion latin datasets: horizontal ICDAR~2013, oriented ICDAR~2015 and curved Total-Text. For comparison see Table~\ref{table:icdar13}, Table~\ref{table:icdar15} and Table~\ref{table:tt}.

Experiments show that our model achieves competitive performance results on all considered datasets. YAMTS establishes new state-of-the-art results in some cases if we take input image resolution into account. It reaches 74.1\% End-to-end recognition metric on ICDAR~2015 using generic lexicon and 73.7\% End-to-end recognition metric on Total-Text dataset, while input images are resized to 1280x768 resolution.

Having in mind the idea that the text recognition head might be undertrained due to complex training pipeline, we freeze all layers except layers that are related to the text recognition head and fine-tune the model during 130K iterations. This fine-tuning does not improve results on ICDAR~2015, but improves quality results upto 74.5\% End-to-end recognition metric on Total-Text dataset, making quality metrics values on par with Mask TextSpotter v3 fine-tuned on TextOCR dataset (\cite{textocr}), see Table~\ref{table:tt} for details.

We also computed End-to-end recognition metric on Open Images V5 validation subset, results are shown in Table~\ref{table:oi}.

In order to understand the effect of using the Open Images V5 Text Annotation, we exclude it from training set and adjust number of epochs so that number of iterations is almost the same as in original training. Obviously we get worse quality metric values of trained model. See details in aforementioned tables.

\begin{table}[H]
\centering 
\begin{threeparttable}
\caption{Results on ICDAR~2013. ‘S’, ‘W’ and ‘G’ mean recognition with strong, weak and generic lexicon, respectively.} 
\begin{tabular*}{\textwidth}{l @{\extracolsep{\fill}} cccccc} 
\hline 
\multirow{2}{*}{Method} & \multicolumn{3}{c}{Word Spotting} & \multicolumn{3}{c}{End-to-end recognition} \\\cline{2-7} %
& S & W & G & S & W & G \\
\hline 
TextBoxes++ & \textbf{96.0} & 95.0 & 87.0 & 93.0 & 92.0 & 85.0 \\
CRAFTS (L-1280) & - & - & - & \textbf{94.2} & \textbf{93.8} & \textbf{92.2} \textsuperscript{\textdagger} \\
MANGO* (L-1440) & 92.9 & 92.7 & 88.3 & 93.4 & 92.3 & 88.7 \\ [1ex]
\hline
YAMTS* (1280x768) & 95.2 & \textbf{95.0} & \textbf{93.3} & 93.6 & 93.3 & 91.0 \\
YAMTS*\textsuperscript{\ding{118}} (1280x768) & 94.1 & 93.1 & 89.8 & 92.1 & 90.8 & 87.0 \\ [1ex]
\hline 
\end{tabular*}
\label{table:icdar13} 
\begin{tablenotes}
\item[\textdagger] means that generic evaluation is performed without the generic vocabulary set.
\item[*] means that the method uses the specific lexicons from \cite{MTSv2}.
\item[\ding{118}] means that the model is trained without Open Images V5 Text Annotation.
\end{tablenotes}
\end{threeparttable}
\end{table}

\begin{table}[H]
\centering 
\begin{threeparttable}
\caption{Results on ICDAR~2015. ‘S’, ‘W’ and ‘G’ mean recognition with strong, weak and generic lexicon, respectively.} 
\begin{tabular*}{\textwidth}{l @{\extracolsep{\fill}} cccccc} %
\hline 
\multirow{2}{*}{Method} & \multicolumn{3}{c}{Word Spotting} & \multicolumn{3}{c}{End-to-end recognition} \\\cline{2-7} %
& S & W & G & S & W & G \\
\hline 
TextBoxes++ & 76.5 & 69.0 & 54.4 & 73.3 & 65.9 & 51.9 \\
Mask TextSpotter v3* (S-1440) & 83.1 & 79.1 & 75.1 & 83.3 & 78.1 & 74.2 \\
CRAFTS (2560x1440) & - & - & - & 83.1 & \textbf{82.1} & 74.9\textsuperscript{\textdagger} \\
MANGO* (L-1800) & 85.2 & 81.1 & 74.6 & 85.4 & 80.1 & 73.9 \\ [1ex]
\hline
YAMTS* (1280x768) & 85.3 & 81.9 & 76.6 & 83.8 & 79.2 & 74.1 \\
YAMTS* (1600x960) & \textbf{87.0} & \textbf{83.6} & \textbf{78.9} & \textbf{85.5} & 80.7 & \textbf{76.1} \\
YAMTS*\textsuperscript{\ding{118}} (1280x768) & 84.4 & 79.9 & 73.4 & 82.8 & 77.2 & 70.5 \\
YAMTS*\textsuperscript{\ding{118}} (1600x960) & 86.8 & 82.4 & 76.7 & 85.3 & 79.8 & 74.0 \\ [1ex]
\hline 
\end{tabular*}
\label{table:icdar15} 
\begin{tablenotes}
\item[\textdagger] means that generic evaluation is performed without the generic vocabulary set.
\item[*] means that the method uses the specific lexicons from \cite{MTSv2}.
\item[\ding{118}] means that the model is trained without Open Images V5 Text Annotation.
\end{tablenotes}
\end{threeparttable}
\end{table}

\begin{table}[H]
\centering 
\begin{threeparttable}
\caption{Results on Total-Text validation. ``None'' refers to recognition without any lexicon. ``Full'' lexicon contains all words in test set.} 
\begin{tabular*}{\textwidth}{l @{\extracolsep{\fill}} cc} %
\hline 
\multirow{2}{*}{Method} & \multicolumn{2}{c}{End-to-end recognition} \\\cline{2-3} %
& None & Full \\
\hline 
ABCNet (Multi-Scale) & 69.5 & 78.4 \\
Mask TextSpotter v3* (S-1000, \cite{MTSv3}) & 71.2 & 78.4 \\
Mask TextSpotter v3* (S-1000, \cite{textocr}) & 74.5 & 81.6 \\
MANGO (L-1600) & 72.9 & \textbf{83.6} \\
CRAFTS (L-1920) & \textbf{78.7} & - \\  [1ex]
\hline 
YAMTS* (1280x768) & 73.7 & 80.1 \\
YAMTS*\textsuperscript{\textdaggerdbl} (1280x768) & 74.5 & 81.5 \\  
YAMTS*\textsuperscript{\ding{118}} (1280x768) & 71.1 & 78.4 \\ [1ex] 
\hline 
\end{tabular*}
\label{table:tt} 
\begin{tablenotes}
\item[*] means that the method uses the specific lexicons from \cite{MTSv2}.
\item[\textdaggerdbl] means that the text recognition head is fine-tuned and the rest layers are frozen.
\item[\ding{118}] means that the model is trained without Open Images V5 Text Annotation.
\end{tablenotes}
\end{threeparttable}
\end{table}

\begin{table}[H]
\centering 
\begin{threeparttable}
\caption{Results on Open Images V5 validation. No lexicon is used.} 
\begin{tabular*}{\textwidth}{l @{\extracolsep{\fill}} cc} %
\hline 
Method & Word Spotting & End-to-end recognition \\ %
\hline 
YAMTS (1280x768) & 58.8 & 51.6 \\
YAMTS (1600x960) & 63.5 & 56.3 \\ [1ex]

\hline 
\end{tabular*}
\label{table:oi} 
\end{threeparttable}
\end{table}

\section{Conclusion}

In this paper we introduced the largest publicly available manually created text annotation for Open Images V5 dataset. We share the annotation to facilitate research in this area and make a further step towards robust and reliable text spotting approaches. Also we presented a simple Mask-RCNN based model architecture with additional text recognition head. The model was trained on dataset including our annotation and showed competitive and even new state-of-the-art performance among certain datasets.

\bibliography{bibliography}

\end{document}